\documentclass[letterpaper, 10 pt, conference]{ieeeconf}  

\usepackage[letterpaper,bindingoffset=0.2in,%
            left=0.667in,right=0.667in,top=0.792in,bottom=0.792in]{geometry}

\IEEEoverridecommandlockouts                              

\usepackage[backend=biber,style=numeric, sorting = none]{biblatex}
\usepackage{graphicx}
\usepackage[ruled,vlined]{algorithm2e}
\usepackage{float}
\usepackage{tabularx}
\usepackage{makecell}
\usepackage{xr-hyper}
\usepackage{xcolor}
\usepackage{hyperref}

\usepackage{array,multirow}

\title{\LARGE \bf
Design and Characterization of a 3D-printed Pneumatically-driven Bistable Valve with Tunable Characteristics
}

\author{Sihan Wang$^{1}$, Liang He$^{1*}$ and Perla Maiolino$^{1}$
\thanks{*This work was supported by the Engineering and Physical Sciences Research Council (EPSRC) Grant EP/V000748/1; S. Wang was supported by the CSC-PAG Oxford Scholarship for his DPhil study.}
\thanks{$^{1}$Authors are with the Oxford Robotics Institute, University of Oxford, UK; \
*Corresponding Author: Liang He
        {\tt\small sihan.wang/liang.he/perla.maiolino@eng.ox.ac.uk}}
}

\begin{document}

\maketitle
\thispagestyle{empty}
\pagestyle{empty}

\begin{abstract}
\textcolor{black}{Although research studies in pneumatic soft robots develop rapidly, most pneumatic actuators are still controlled by rigid valves and conventional electronics. The existence of these rigid, electronic components sacrifices the compliance and adaptability of soft robots.} Current electronics-free valve designs based on soft materials are facing challenges in behaviour consistency, design flexibility, and fabrication complexity. Taking advantages of soft material 3D printing, this paper presents a new design of a bi-stable pneumatic valve, which utilises two soft, pneumatically-driven, and symmetrically-oriented conical shells with structural bistability to stabilise and regulate the airflow. The critical pressure required to operate the valve can be adjusted by changing the design features of the soft bi-stable structure. Multi-material printing simplifies the valve fabrication, enhances the flexibility in design feature optimisations, and improves the system repeatability. In this work, both a theoretical model and physical experiments are introduced to examine the relationships between the critical operating pressure and the key design features. Results with valve characteristic tuning via material stiffness changing show better effectiveness compared to the change of geometry design features (demonstrated largest tunable critical pressure range from $15.3$ to $65.2$ kPa and fastest response time $\leq 1.8 s$).

\end{abstract}

\section{INTRODUCTION}

Pneumatic soft robots are of interest to an increasing number of researchers due to their compatibility in highly unstructured and dynamic environments \cite{lee2017soft}. Most soft pneumatic robots are based on the interaction between regulated air and soft elastomeric materials with high adaptability and compliance. This mechanism brings various advantages including (i) low cost; (ii) flexible manufacture choices such as \textcolor{black}{silicone} casting, 3D printing, and blow film extrusion; (iii) safe human-robot interaction; and (iv) adaptability to environments or objects with unknown kinematics or dynamics. These valuable characteristics lead to the development of various pneumatic actuators with different structures, action types, pressure ranges, and fabrication methods \cite{walker2020soft}. However, most soft pneumatic actuators still require conventional electronic controllers and hard valves to regulate the internal fluid and control the actuation, sacrificing some benefits of being soft \textcolor{black}{\cite{tanaka2021dynamic,he2021method,tang2020leveraging}}. The existence of these tethered external rigid pressure systems severely limits the mobility and application fields of soft robots. This is particularly challenging in the design of soft robots for extreme environment operations (e.g., nuclear plant exploration with high-radiation, tissue interaction during metal-free scenarios in magnetic resonance imaging). Although rigid electronic controllers allow multimodal sensor integration into the system \cite{he2020soft}, the conversion of sensory information into an electrical signal before being processed to pneumatic signal control brings additional system complexity and limits the application in extreme conditions. 
\begin{figure}[!t]
\begin{center}
  \includegraphics[width=0.95\columnwidth]{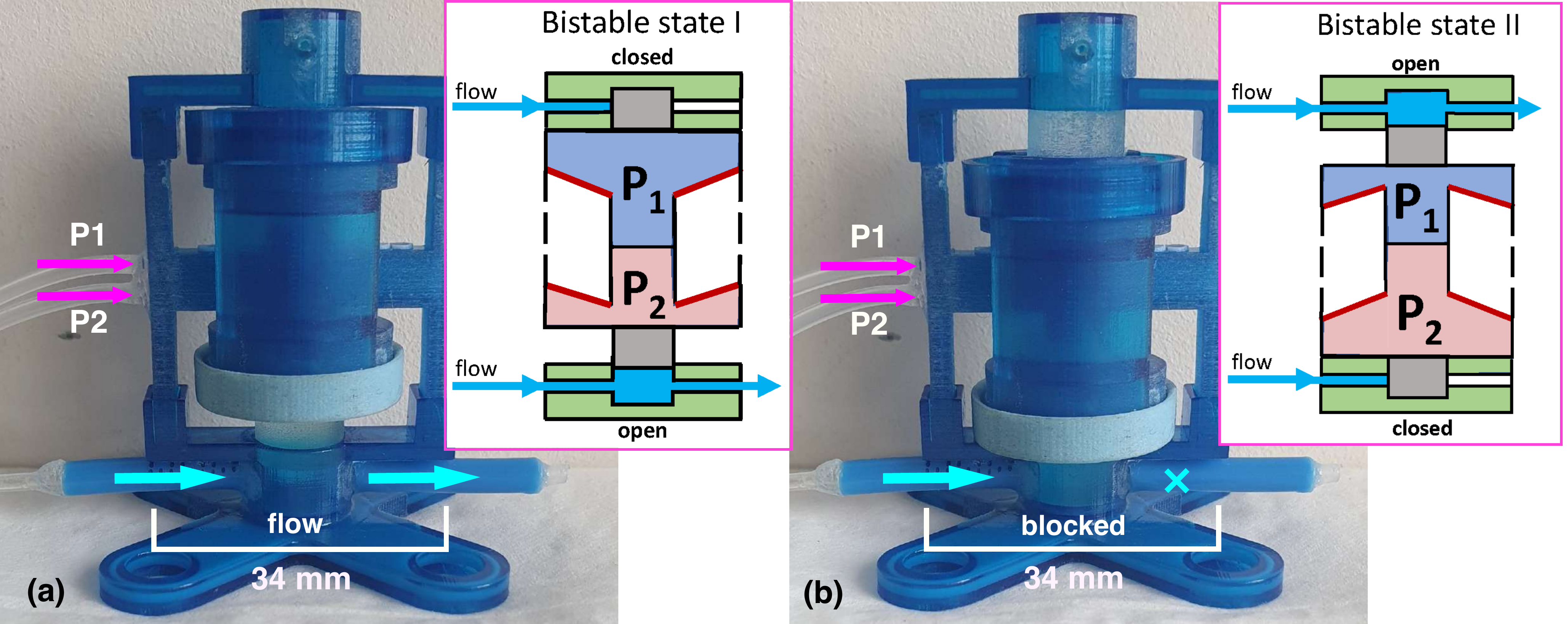}
  \vspace{-3mm} 
  \caption{Pneumatically-driven bistable valve at each bistable state \textbf{(a)} and \textbf{(b)}. $P_1$ and $P_2$ represent the two pressures that are used to switch the state of the valve.}
  \label{real}
  \vspace{-5mm}
  \end{center}
\end{figure}

A potential solution is to replace the rigid pneumatic power transmission with soft computational elements, where the adaptability and usability of the soft robotic system can be improved. The challenge for designing an ideal computational system for soft pneumatic robots is how to store information in the physical structure without a continuous external energy supply while receiving/processing pneumatic signals via soft interfaces. For instance, a soft robot with integrated soft computation elements should perform on-board computation without converting the pneumatic information to electronic signals. The properties and behaviours of the transmission devices should also come with high design flexibility so that the individual control element can be adapted into different applications. The development of electronic-free valves is a critical contribution to the future of fully soft robots required in extreme environments. 

 
A bi-stable structure allows the electronic valve to be operated without a continuous energy supply. Each stable state can be used as either ``blocking" or ``releasing" the flow. The use of the semi-spherical bistable structure in soft-matter computing has previously been demonstrated in \cite{rothemund2018soft}. A passive bi-stable membrane driven by external pressure is used to press and make the internal tubing buckle, therefore regulating the flow based on the input pneumatic signal. The valve successfully achieves the flow regulation based on soft materials without requiring continuous energy supply. However, the \textcolor{black}{silicone} casting process makes the fabrication time-consuming, while the key design dimensions are less flexible to adjust. The system repeatability is also low as it requires a large amount of manual work during fabrication.
\cite{drotman2021electronics} used pneumatic circuits composed of a similar semi-spherical bistable structure to control the basic walking gaits of a soft legged robot without electronics. These circuits generate oscillatory signals, which are directly used to drive the pneumatic actuators, therefore successfully controlling the locomotion of the legged robots. The idea of buckling a soft tube as a mechanical ``switch" to block the airflow is easy for soft robot integration \cite{luo2019soft}. However, the behaviour consistency during operation is normally of concern when the system complexity increases with more transmission units. The kinking of the tube also brings large local stress in the elastomer, which reduces the durability of the structure and decreases the maximum controllable pressure.

A different approach for electronics-free pneumatic control is developed by \cite{9044752}, where the interaction between three different pneumatic actuators in a soft crawler is controlled by three
soft switch-valves adhered to the skins of the robot. The dominant advantage of this work is that, the parameters of these valves can be adjusted from the exterior of the robot through simple geometrical changes. However, these valves work with only pneumatic actuators with large working stroke, as it requires sufficient deformation to kink the elastomeric tubes and block the flow.  Soft material 3D printing technique \cite{gul20183d,he2021method} provided another possible approach to create bi-stable structures with enhanced model repeatability, increased design flexibility, and ease in fabrication. Compared with \textcolor{black}{silicone} casting, 3D printing of thin bi-stable membrane brings reliable consistency in each batch of fabrication, which guarantees the repeatability of the snap-through behaviour. High design flexibility is also introduced to the valve development since the material stiffness and membrane geometry can be effectively tuned to change the required control conditions. 

Thus, this work presents the novel design of a 3D printed electronic-free valve which takes pneumatic signal as input to regulate the airflow in specific air channels. The valve design consists of two soft, pneumatically-driven, and symmetrically-oriented conical shells with structural bistability and soft pistons to achieve stable regulation of the airflow. The paper discusses how the valve behaviour (critical switching pressure and response time) can be altered by changing several key design dimensions (material stiffness, structure geometry) so that it is ready for immediate inclusion in a wide range of pneumatic systems. This work also investigates how different support removal techniques affect the valve behaviour and durability.




\section{Design and fabrication of the bistable valve with tunable characteristics}
\subsection{Working principle of the bistable valve}\label{sec_working_principle}

The operation of the valve makes use of a bistable structure with two internal chambers to get rid of the requirement of a continuous power supply. The bistable structure has two stable states and can be switched from one state to another by applying a pneumatic signal (positive pressure) to one of the chambers. The state of the bistable structure is used to decide whether or not the controlled air channel should be blocked. The blocking is achieved by two pistons attached to the bistable structure. \textcolor{black}{When any of the pistons is pressed into the piston cylinder, the air cannot go through the corresponding air channel any more as the piston is physically in its way. The high accuracy of the Polyjet 3D printer used in this work enables optimal fit between the piston and the cylinder to prevent leakage.}

\begin{figure}[!t]
\begin{center}
  \includegraphics[width=0.85\columnwidth]{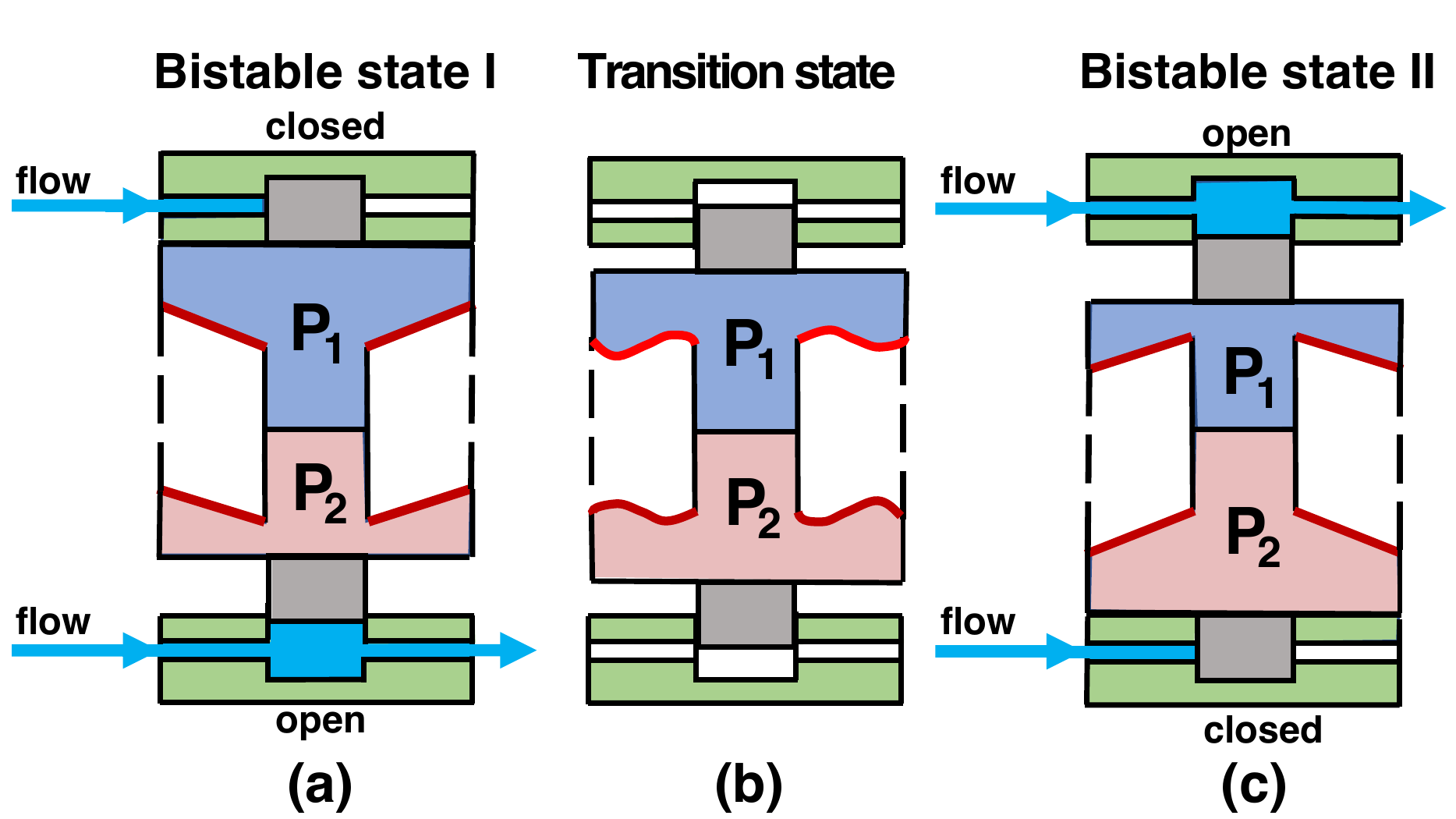}
  \vspace{-3mm} 
  \caption{Schematics of the bistable valve. \textcolor{black}{The valve consists of two symmetrical compliant bistable conical shells (shown by red lines), switching in the opposite state (one ``expanded" and one ``contracted").} The internal pressures in each shell ($P_1$, $P_2$) indicate the valve control signal. \textbf{(a)} Bistable state I: the top chamber is ``expanded" while the bottom chamber is ``contracted." This means the top air channel is closed while the bottom air channel is opened. \textbf{(b)} An unstable transition state located between the two stable states. \textbf{(c)} Bistable state II: the top chamber is ``contracted" while the bottom chamber is ``expanded." This means the bottom air channel is closed while the top air channel is opened.}
  \label{Fig_1_design}
  \vspace{-5mm} 
  \end{center}
\end{figure}

One typical switching stroke for the valve is shown as followed. (i) the valve remains stable in the state shown in Fig.~\ref{Fig_1_design}(a). In this case, the top control channel is blocked while the bottom channel is released. (ii) Once sufficient positive pressure is applied to the bottom shell, the bottom shell will be switched to an ``expanded" state. Due to the restricted length of the tether (black dashed line in Fig.~\ref{Fig_1_design}(a)), only one shell can be in the ``expanded" state at one time, which means the top shell will be pulled to its ``contracted" state. The top controlled air channel is then unblocked, while the bottom controlled channel is blocked as shown in Fig.~\ref{Fig_1_design}(b). (iii) The driving pressure in the bottom shell is then removed. The state of both control channel will not change thanks to the bistability of the structure.

The snap-through process of the soft bistable structure determines the switching behaviour of the valve. This means by adjusting the design of those bistable conical shells, the critical pressure required to switch the state of the valve can be altered to fit the users' needs. 

\subsection{Fabrication}
\label{fabrication}
The use of a multi-material 3D printer (J735, Stratasys Ltd, USA) allows the stiffness of the bistable valve to be digitally tuned in the design stage with precision control of the mechanical property. The J735 uses Polyjet technology, which deposits multiple droplets of uncured material, rolls them to combine and then UV cures the blend. \textcolor{black}{By combining VeroWhite (a rigid plastic-like material with a quoted tensile strength of 50-65MPa and Shore hardness of 83-86D) and Agilus30 (a soft and rubber-like material with a quoted tensile strength of 2.1-2.6MPa and Shore hardness of 30A), a finite set of digital materials with variable stiffness can be printed (40, 50, 60, and 70 shore hardness). The digital material creation is made with the ``digital material" function in the GrabCAD Print slicer (Stratasys, Ltd. USA) by selecting the shore hardness, while the repeatability has been well characterized by several previous works \cite{akbari2018enhanced,slesarenko2018towards}.} These digital materials with tunable hardness are used to print the compliant conical shell (shown in red in Fig.~\ref{Fig_1_design}). The piston is printed by pure Agilus 30, so its compliance ensures reliable sealing. Two different support removal techniques (physical removal and chemical bathing) are investigated and their effect on the valves are compared and discussed. Physical removal is achieved by using a water-jetting machine (GEMINI Cleaning ECO-400) while chemical removal is achieved by placing the valve in a tank (GEMINI SSR-550) filled with chemical support removal solution (0.02 kg/L Sodium Hydroxide and 0.01 kg/L Sodium Metasilicate).

 \section{Theoretical Model}
\label{sec_theoretical_model}
\begin{figure}[!t]
\begin{center}
  \includegraphics[width=\columnwidth]{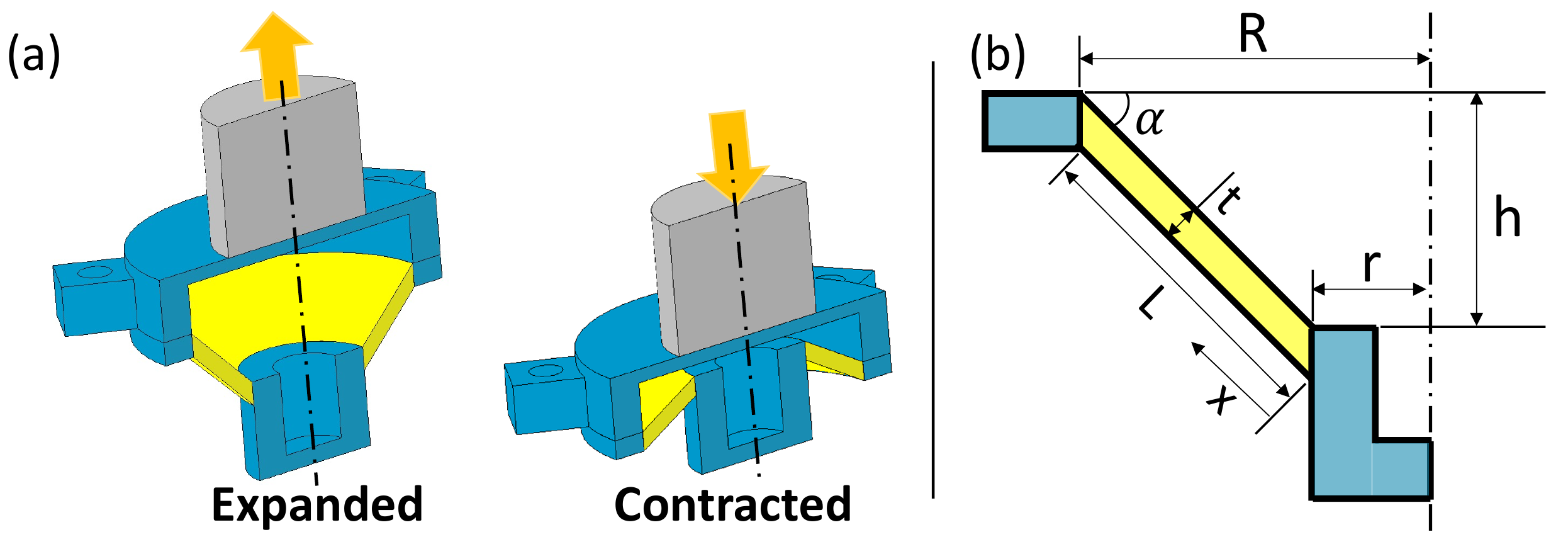}
  \vspace{-3mm} 
  \caption{\textbf{(a)} Cross-sectional view of one bistable shell under both ``expanded" and ``contracted" state. The blue part is printed with Stratasys Vero, the grey piston at each end is printed with Stratasys Agilus 30, while the yellow part is printed with digital material which is a combination of Stratasys Vero and Agilus 30. \textbf{(b)} Annotated cross-sectional view of the bistable structure used in the valve (yellow region). $R$ is the outer radius of the soft conical shell. $r$ is the inner radius of the soft conical shell. $h$ is the height of the conical shell (the vertical displacement between the outer and inner edge). $t$ is the thickness of the conical shell, $x$ is the distance along the cone slant counting from the inner edge, and $\alpha$ is the slope angle of the conical shell. }
  \label{Fig_2_model}
  \vspace{-5mm}
  \end{center}
\end{figure}

 Fig.~\ref{Fig_2_model}(a) shows the cross-sectional view of one bistable shell under both ``expanded" and ``contracted" state. The state can be switched by the internal pressure as it exerts a uniformly distributed load to the entire internal surface of the shell, therefore expanding or shrinking the internal volume. The volume change are mostly contributed by the deformation of the conical shell (shown in yellow in Fig.~\ref{Fig_2_model}) as it is printed with more flexible material combined with Agilus30 and VeroWhite. When the structure deforms from one state to another, the height of the cone $h$ decreases to zero and then increases on the opposite direction, therefore the length of the cone slant $L$ decreases and then increases again due to the Pythagoras' theorem. This shortening process compresses and buckles the conical shell and therefore brings bistability to the structure.

To further investigate the feasibility and snap-through behaviour of this pneumatically driven bi-stable structure, a theoretical model is built and investigated. The relationship between the elastic strain energy curve stored in the conical shell and the structural deformation (in terms of the height of the cone $h$) is of specific interest, as the local minimum points on the strain energy curve tell the possible stable states and how much deformation is required to reach those states. After locating each stable state, by equating the mechanical work done by the compressed air in the chamber and the elastic strain energy stored in the structure, the relationship between the driving pressure and the structure deformation can be derived. The obtained pressure-displacement relationship can then be analysed to obtain the critical pressure required to switch the structure from one state to another.\\

For model simplification, the conical shell is sliced into infinitely small sector with central angle $\delta \theta$ (see Fig. \ref{Fig_2_model}), where the slant surface in each slice are treated as a beam with varying cross-sectional area. Several assumptions are made during the consideration of this model: (i) Only the deformation in slant surface is considered. All other parts contribute zero internal volume change and store zero elastic strain energy; (ii) The material follows a linear and homogeneous model; (iii) The deformation process is treated as quasi-static.

Based on to the assumptions listed above, the slant undergoes an axial compression when the bistable state is switched. The amount of that compression can be derived from the Pythagoras' theorem: 
\begin{equation}
    \Delta l = L-\sqrt{(R-r)^2+h^2}
    \label{pythagoras}%
\end{equation}
where $\Delta l$ is the amount of compression in this beam, $L$ is the original slant length of the conical shell, $R$ is the radius of the outer ring, $r$ is the radius of the inner ring and $h$ is the vertical displacement between the two rings.

When the amount of compression is not large enough to buckle the slant, there's only deformation due to the negative axial extension. The elastic energy due to this axial extension is given by \cite{megson2019structural}:
\begin{equation}
    \textcolor{black}{U_{axial}}= \int_{0}^{L} \frac{F^2}{2EA(x)} dx
    \label{axial energy from force}%
\end{equation}
\begin{equation}
    \textcolor{black}{A(x) = \int_0^{2\pi}(r+x\cdot cos\alpha) t\cdot d \theta 
    \label{a_x}}
\end{equation}
where $\textcolor{black}{U_{axial}}$ is the elastic energy stored due to the axial extension, $x$ is the distance along the slant beam starting from the inner end, $E$ is the Young's modulus of the material, $A(x)$ is the cross-sectional area of the slant beam, $\alpha$ is the slant angle made with the horizontal line, $t$ is the shell thickness and $F$ is the axial compressive force within the slant beam. The relationship between this compressive force $F$ and beam overall compression $\Delta l$ is:
\begin{equation}
    \Delta l = \int_0^L \frac{F}{EA(x)} dx
    \label{deltal from force}%
\end{equation}

As the displacement of the structure keeps increasing, the amount of compression will eventually reach a certain point where buckling occurs. To find the exact amount of displacement required for triggering the buckling, Rayleigh-Ritz \cite{megson2019structural} method is used. Rayleigh-Ritz method is a energy-based method, stating that the critical point for buckling occurs when the mechanical work done by the compressive force equals the strain energy stored in the structure due to bending just after the buckling. The details about this statement are represented by Equation \ref{Raleigh model}.

\begin{equation}
    \int_{0}^{L} \frac{1}{2} EI(x)(\frac{d^2w}{dx^2})^2dx = F_c \times \Delta l 
    \label{Raleigh model}%
\end{equation}
\textcolor{black}{
\begin{equation}
    I(x) =\int ^{2\pi}_0 \frac{1}{12}(r+x\cdot cos\alpha)t^3 d\theta
    \label{I_x}
\end{equation}}
where $I(x)$ is the second moment of area of the slant beam, $w$ is the local deflection of the beam under bending and $F_c$ is the critical compressive force within the beam at the point of buckling.

In order to apply Equation \ref{Raleigh model} into use, it is required to assume a buckling field $w(x)$. Our design comes with two thin-panel hinges located at both ends of the slant, these hinges undergo large rotation without causing too much bending moment. Therefore, the slant can be treated as a pin-ended structure, which has the following buckling shape:
\begin{equation}
    w(x) = C[sin(\frac{\pi x}{L})]
    \label{buckling shape assumption}%
\end{equation}

This buckling shape is chosen as it brings zero bending moment at both ends. \textcolor{black}{ The $C$ value in equation \ref{buckling shape assumption} as a function of the overall beam compression $\delta l$ can be found by solving Equation \ref{continuous pythagoras}. Equation \ref{continuous pythagoras} is derived by considering Pythagoras' theorem at each infinitely small segment of the beam and integrate the incremental axial extension over the entire beam length.}
\begin{equation}
    \Delta l  = \int_0^L\frac{1}{2}(\frac{dw}{dx})^2dx
    \label{continuous pythagoras}%
\end{equation}

By substituting Equation \ref{continuous pythagoras} into Equation \ref{Raleigh model}, the buckling load can be found by the following equation:
\begin{equation}
    F_c = \frac{\int_{0}^{L} \frac{1}{2} EI(x)(\frac{d^2w}{dx^2})^2dx}{\int_0^L\frac{1}{2}(\frac{dw}{dx})^2dx}
    \label{critical compressive force}%
\end{equation}

After obtaining the critical axial load in the slant at the point of buckling, the corresponding deformation can then be obtained by:
\begin{equation}
    \Delta l_c = \int_0^L \epsilon_c(x) dx = \int_0^L \frac{F_c}{EA(x)}dx
    \label{critical deltal}
\end{equation}
where $\Delta l_c$ is the amount of compression exhibited by the beam at the critical point of buckling. By substituting $\Delta l_c$ into equation \ref{pythagoras}, the critical vertical displacement at the point of buckling $h_c$ can be obtained.

The elastic energy stored in slant after the buckling is due to the bending caused by the buckling shape. The amount of that elastic energy is obtained by Equation \ref{Raleigh model}.

All equations stated above are sent into Matlab Symbolic Toolbox, and the relationship between the elastic strain energy and the structural deformation $\textcolor{black}{U(h)}$ can be obtained. When the structure is out of the calculated buckling range, the strain energy depends on the axial energy  expressed in Equation \ref{axial energy from force}. \textcolor{black}{When the structure is within its buckling range (the amount of compression in the slant is larger than the threshold value), the strain energy is the sum of the energy due to bending (see Equation \ref{Raleigh model}) and the accumulated axial elastic energy during the axial compression stage before buckling (the value of $\textcolor{black}{U_{axial}}$ in Equation \ref{axial energy from force} when $F = F_c$). It is assumed that the tether connecting the two bistable shells have no deformation. This means the two shells come with the same elastic strain energy during the operation of the valve. Therefore, the total elastic strain energy is obtained by doubling the elastic strain energy of one single shell.}

The relationship between driving pressure and structural deformation $p(h)$ is obtained based on the conservation of energy.That means the work done by the compressed air in the chamber equals the elastic strain energy stored in the structure, which leads to:
\begin{equation}
    \int p(h) dV = U(h)
    \label{energy conservation}%
\end{equation}
where $V(h)$ is the volume within the shell and $p(h)$ is the driving pressure required for different structure displacement. By taking differentiation of both sides in Equation \ref{energy conservation}, we have:
\begin{equation}
    p(h) = \frac{\frac{dU(h)}{dh}}{\frac{dV(h)}{dh}}
    \label{energy conservation 2}
\end{equation}

\begin{figure}[!t]
\begin{center}
  \includegraphics[width=0.75\columnwidth]{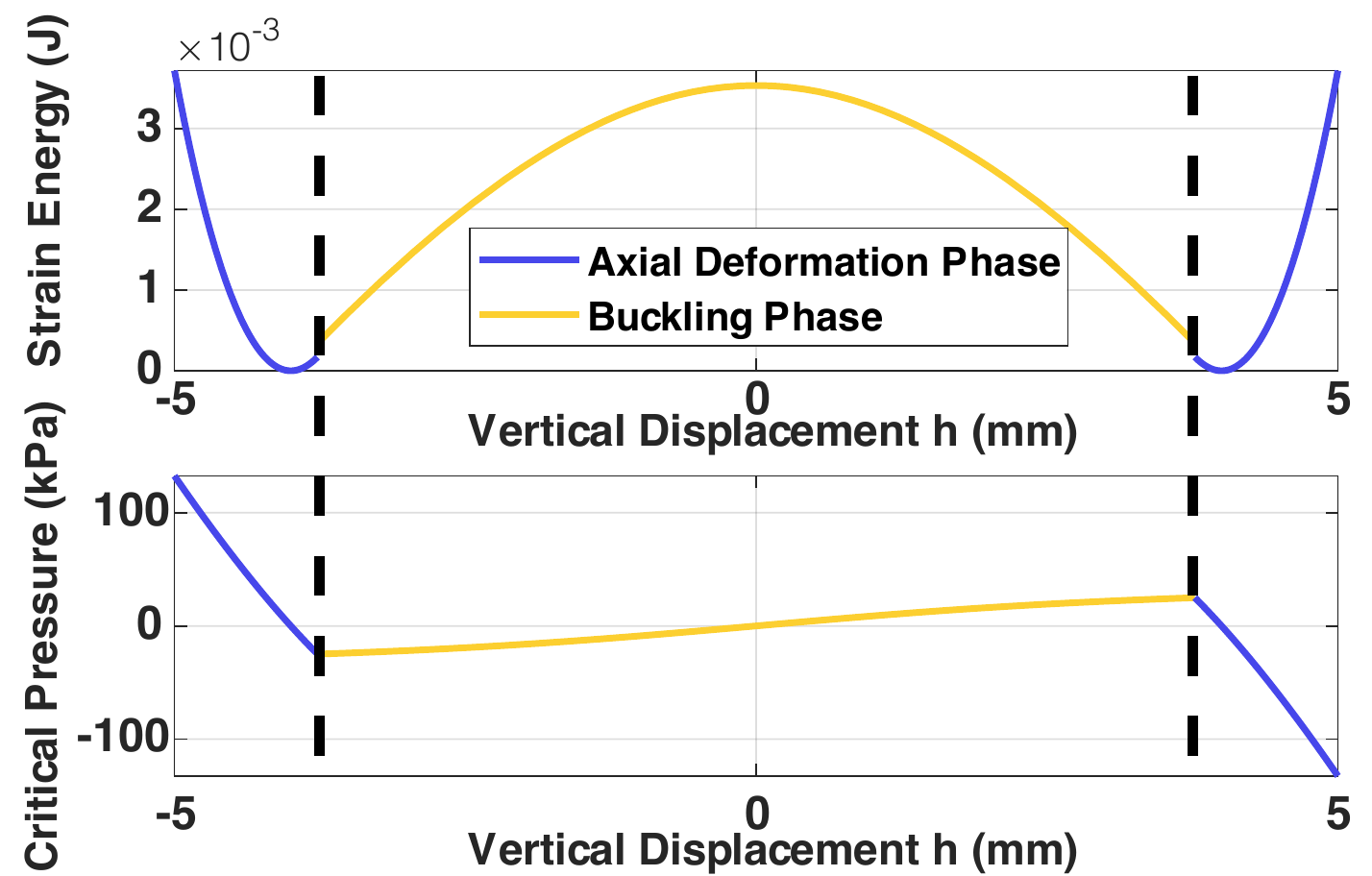}
  \vspace{-3mm} 
  \caption{The elastic strain energy stored $U(h)$ and the corresponding driving pressure $p(h)$ over structural deformation. The deformation is represented by the vertical displacement between the outer and inner edge of the conical shell, shown as $h$ in Fig.~\ref{Fig_2_model}. \textcolor{black}{In this figure, the membrane material is chosen as 50A shore hardness. The membrane thickness is 1mm and the slope angle is 45$^o$. The Young's modulus used for this digital material is 1.65 MPa.}}
  \label{Fig_3_model_result}
  \vspace{-5mm}
  \end{center}
\end{figure}

Fig. \ref{Fig_3_model_result} shows a typical strain energy curve and pressure-displacement curve. Two local minimum points are located on the strain energy curve, while each of them contribute to one possible bistable state. The critical pressure required to switch from one state to another is determined by the largest pressure value observed on the pressure-displacement curve within those two bi-stable states.

\section{Experimental setup and protocols}

In this study, critical switching pressure is evaluated to reflect the tunable characteristic of the bi-stable valve. The critical switching pressure is defined as the minimum pressure difference between two valve chambers required to switch the valve state. The valve response time is also measured as the time taken for the valve to complete the state switching process. The goal of the experimental characterization is to find out the critical pressure and response time of the valves under different design parameters, including the shell material hardness, the conical shell thickness, and the shell slope angle.


\subsection{Experimental Setup and Basic Procedure}
\begin{figure}[!t]
\begin{center}
  \includegraphics[width=\columnwidth]{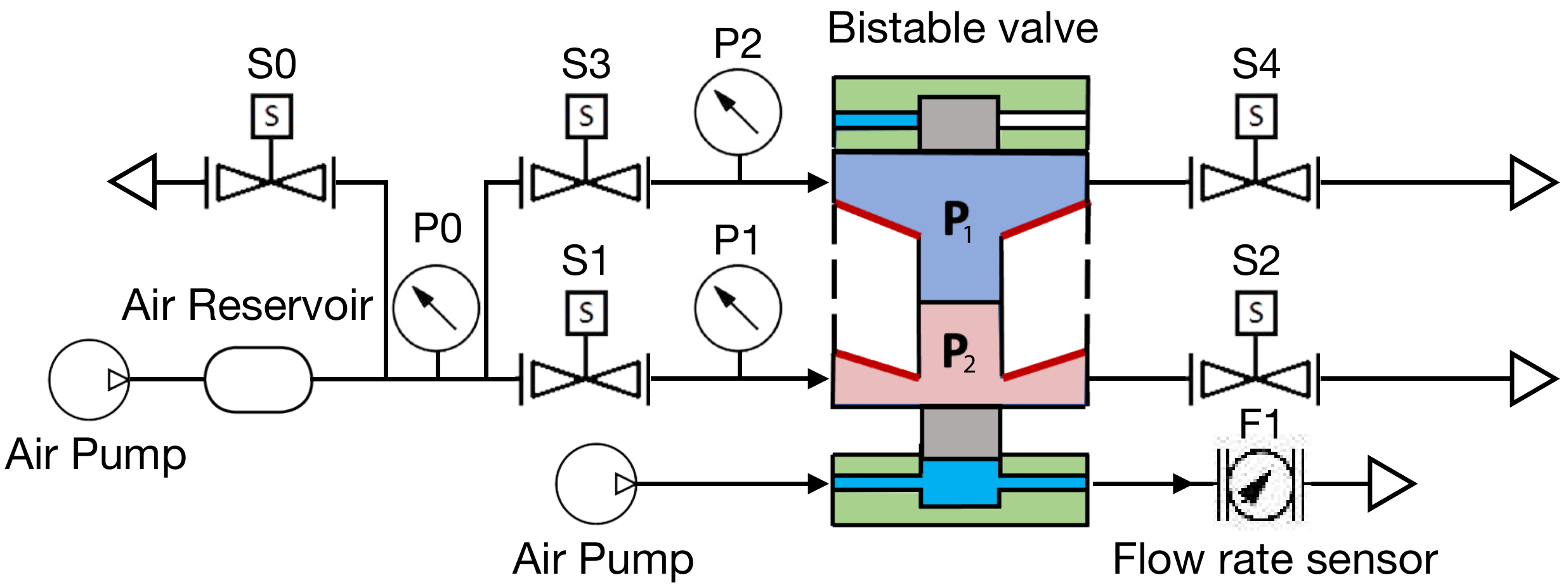}
  \vspace{-3mm} 
  \caption{Schematics of the pneumatic system used for characterization. P0, P1 and P2 stand for pressure sensor 0,1 and 2. S0,S1,S2,S3,S4 stand for solenoid valve 0,1,2,3 and 4. F1 represents the flowrate sensor. For valve activation, the two chambers of the bistable valve ($P_1$,$P_2$) are independently connected to the pressure system to achieve rapid pressure engagement and venting. To detect whether the air channel is blocked or not by the bistable valve, the controlled air channel (located at the bottom of the bistable valve) is connected with an air pump and a flow rate sensor.}
  \label{pneumatic_system}
  \vspace{-5mm}
  \end{center}
\end{figure}

The setup of the conventional pump-valve system to perform the characterization is shown in Fig.~\ref{pneumatic_system}. Two displacement pumps (SIMILK, mini air pump, rated airflow of 2 L/min) are used. One is used to set up the required driving pressure to either chamber of the valve, while the other pump is used to apply constant air flow to the controlled air channel. The driving pressure here is defined as the pressure applied to one of the valve chambers during our experiments. While as we stated before, the critical pressure is the minimum driving pressure required to switch the state of the valve. To achieve rapid pressure settling in the pressure chamber, smooth out fluctuation and avoid overshoots, a reservoir (rigid plastic, with the volume of 250 ml) is used. Multiple solenoid valves (VDW10AA, SMC) are used to engage and vent the driving pressure of each chamber. A volume flowrate sensor (MEMS D6F-P Flow Sensor, OMRON) connected in series with the controlled air channel is used to reflect whether the air channel is successfully blocked by the pneumatic signal. Three independent pressure sensors ($\pm$100 kPa, PSE 543-R06, SMC Corporation, Japan)) are used to monitor the pressure in the reservoir and the two valve chambers respectively. Control signals were generated, and data were captured using a data acquisition device (USB-6341, National Instruments) at 10kHz sampling rate.

For sample preparation, five batches of valve samples were fabricated. The sample key dimensions and post-processing technique are presented in Table \ref{tab:experimentalSamples}.

\begin{table}[!t]
\renewcommand{\arraystretch}{1.2}
\footnotesize\sf\centering
\caption{Experimental sample preparation}
\label{tab:experimentalSamples}
\vspace{-5mm} 
\begin{center}
    \begin{tabular}{p{1cm}| p{1cm} |p{1cm} |p{1 cm} |p{1cm} } \hline

\multicolumn{1}{|c|}{\multirow{2}{*}{\thead{Batch\\ (Quantity)}}} & \multicolumn{3}{c|}{Design parameters} & \multicolumn{1}{c|}{\multirow{2}{*}{\thead{Support\\ Removal}}} \\ \cline{2-4}
\multicolumn{1}{|c|}{}                  & \multicolumn{1}{c|}{$S_A$} & \multicolumn{1}{c|}{$t$(mm)} & \multicolumn{1}{c|}{$\alpha$($^o$)} & \multicolumn{1}{c|}{}                  \\ \hline

\hline
\hline
\multicolumn{1}{|c|}{A (5) }                  & \multicolumn{1}{c|}{\thead{\footnotesize{30, 40,}\\ \footnotesize{50, 60,}\\ \footnotesize{70}}} & \multicolumn{1}{c|}{1} & \multicolumn{1}{c|}{45} & \multicolumn{1}{c|}{Chemical}                  \\ \hline

\multicolumn{1}{|c|}{B (7) }                  & \multicolumn{1}{c|}{50} & \multicolumn{1}{c|}{\thead{\footnotesize{0.7, 0.8, 0.9,}\\ \footnotesize{1.0, 1.1, 1.2,}\\\footnotesize{ 1.3}}} & \multicolumn{1}{c|}{45} & \multicolumn{1}{c|}{Chemical}                  \\ \hline

\multicolumn{1}{|c|}{C (7) }                  & \multicolumn{1}{c|}{50} & \multicolumn{1}{c|}{1} & \multicolumn{1}{c|}{\thead{\footnotesize{30, 35, 40,}\\ \footnotesize{45, 50, 55,}\\\footnotesize{ 60}}} & \multicolumn{1}{c|}{Chemical}                  \\ \hline

\multicolumn{1}{|c|}{D (6) }                  & \multicolumn{1}{c|}{50} & \multicolumn{1}{c|}{0.8} & \multicolumn{1}{c|}{45} & \multicolumn{1}{c|}{\thead{3x Chemical\\3x Physical}}                  \\ \hline

\multicolumn{1}{|c|}{E (2) }                  & \multicolumn{1}{c|}{50} & \multicolumn{1}{c|}{0.8} & \multicolumn{1}{c|}{45} & \multicolumn{1}{c|}{\thead{1x Chemical\\1x Physical}}                  \\ \hline

\multicolumn{5}{|p{8cm}|}{*\footnotesize $S_A$ refers to the ASTM D2240 type A hardness, $t$ refers to the thickness of the conical shell. $\alpha$ refers to the slope angle of the conical shell. Chemical support removal refers to chemical bathing. Physical support removal refers to water-jetting. \textcolor{black}{For all samples, the radius of the conical shell outer ring $R$ is 8 mm, and the inner ring radius $r$ is 4mm.} }                 \\ \hline

    \end{tabular}
\end{center}
 \vspace{-3mm}
\end{table}

\begin{algorithm}[!t]
\small
 \KwResult{10kHz sampling data of pressure in reservoir ($P_{reservoir}$), pressure in both valve chambers ($P_{1},P_{2}$) and volume flowrate in the controlled air channel ($Q_{controlled}$).}
 $P_{target}$ $\leftarrow$ 1 kPa;  \\
 $\delta p$ $\leftarrow$ 1 kPa for Batch A,B and C;\\
 $\delta p$ $\leftarrow$ 0.2 kPa for Batch D and E;\\
 \While{$P_{critical}$ not found yet}{
  Valve S1 to 5 $\leftarrow$ OFF;\\
  set $P_{reservoir}$ to $P_{target}$ via a PID controller;\\
  Valve S4 $\leftarrow$ ON ;\\
  Valve S2 $\leftarrow$ ON ;\\ 
  Pause for 10 seconds;\\
  Valve S2 $\leftarrow$ OFF $\;$  Valve S1 $\leftarrow$ ON ;\\
  Pause for 5 seconds;\\
  \eIf{$Q_{controlled}$ $\leq$ 5 mL/min}{
   $P_{critical}$ $\leftarrow$ $P_{target}$;\\
   }{
    $P_{target}$ = $P_{target}$ + $\delta p$;\\
  }
  Valve S1 and 4 $\leftarrow$ ON;
  Pause for 5 seconds;\\
  Valve S1 and 4 $\leftarrow$ OFF;
 }
  Save $P_{reservoir}$,$P_{1},P_{2}$ and $Q_{controlled}$
 \caption{Procedure for measuring the critical pressure of a valve.}
\end{algorithm}

Batch A,B and C aims to find how the valve behaviour (critical pressure and response time) is affected by the three key design dimensions (shell material hardness, shell thickness, shell slope angle) respectively. Batch D aims to investigate the effect of different support-removal techniques on the valve behaviour. It also aims to investigate the consistency between each fabricated valve with the same dimensions. Batch E is used for fatigue tests aiming to examine the durability of the valve. 



\subsection{Experimental protocol of the critical pressure and response time characterization}
\label{sec_experiment_protocol}

For each sample in batch A,B,C and D, the critical pressure and response time were measured. \textcolor{black}{The measurements were taken repeatedly (3 times for batch A,B and C; 5 times for batch D) with 0.5 hrs relaxation between each trial to ensure data reliability.} The critical pressure is found empirically by applying a specific driving pressure ($P_{target}$ in Algorithm 1) to one of the valve chambers and observing whether the valve state is successfully switched. Starting from a randomized initial value which is insufficient to switch the state of the valve, the driving pressure $P_{target}$ is incrementally increased with a small step until the valve state is successfully switched. Note that the pressure in the valve chamber is reset to the atmosphere pressure after each trial to get rid of any hysteresis effect. \textcolor{black}{As the valve comes with a completely symmetrical structure, the switching pressure at both switching direction are assumed to be the same, so the switching direction used during measurement is randomly chosen. The detailed procedure is shown in Algorithm 1.}

\begin{figure}[!t]
\begin{center}
  \includegraphics[width=0.9\columnwidth]{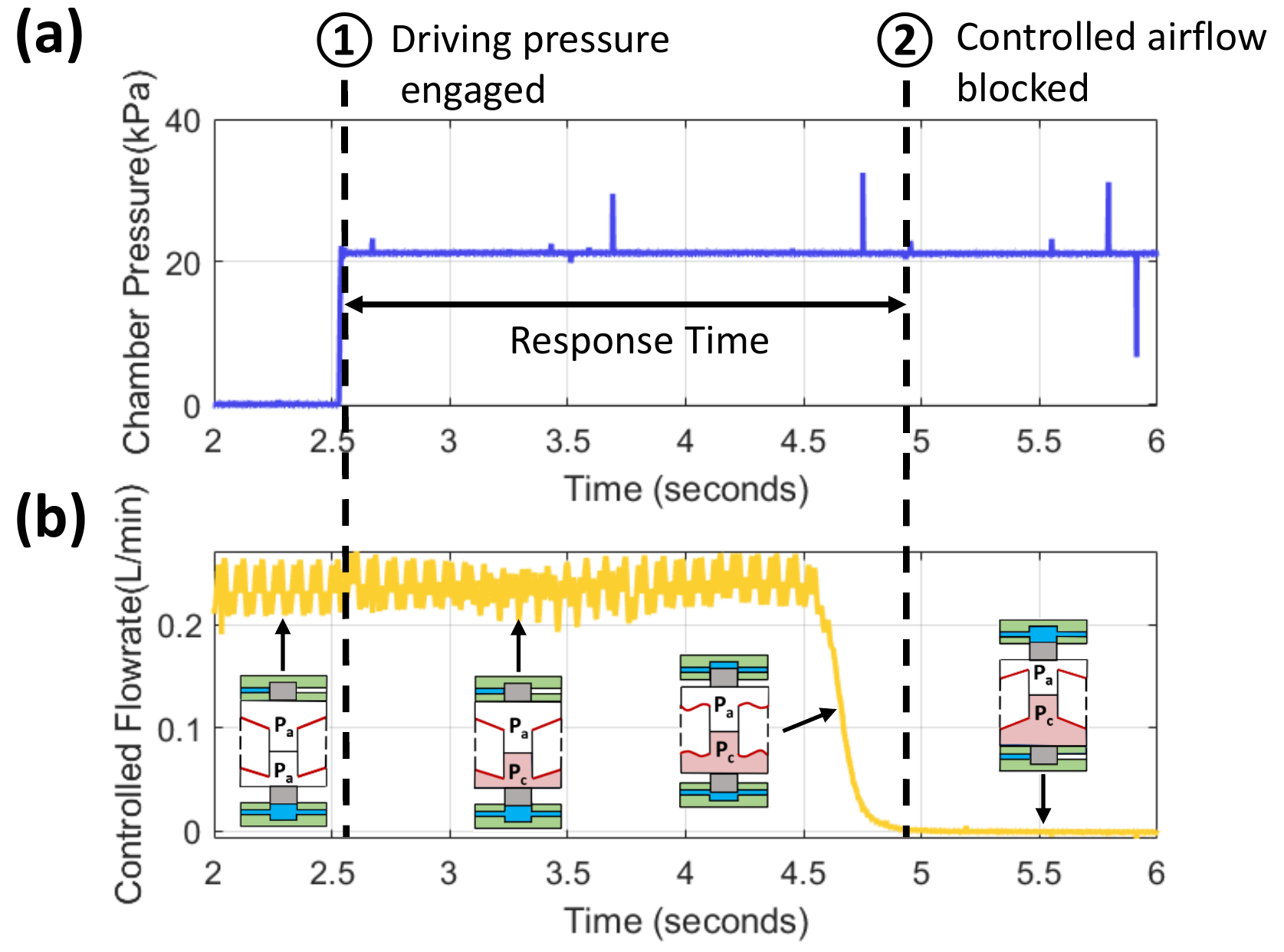}
  \vspace{-3mm} 
  \caption{This demonstrates how the response time is obtained from the time-series data. (a) shows the pressure in the chamber of the valve and (b) shows the volume flow rate of the controlled air channel. $P_a$ refers to the atmosphere pressure while $P_c$ refers to the critical pressure of the valve. ``Driving pressure engaged" is defined when the pressure in the valve chamber reaches and stabilises at the desired level. ``Controlled airflow blocked" is defined when the flow rate controlled reaches and stabilises at a level which is lower than 5mL/min. The response time is obtained by calculating the time difference between these two time stamps.}
  \label{data_processing}
  \end{center}
\end{figure}

The response time of the valve is defined as the total time required by the valve to shut down the controlled air flow after the engagement of that critical switching pressure. This is obtained by post-processing the time-series data obtained from the experiment. Fig.~\ref{data_processing} shows how the response time is obtained from the raw time-series data. 


\subsection{Experimental protocol of the fatigue test}
For each sample in batch E, fatigue tests are performed to examine the durability of the valves. The fatigue tests are preformed by repeatedly switch the valve state from one to another (shown in Algorithm 2) for 500 working cycles. 
To further challenge the reliability of the valves, the driving pressure used in fatigue test is set to 35 kPa, which is larger than the critical switching pressure. 
The response time during each working cycle is recorded. Meanwhile, the critical pressure before and after these fatigue tests are measured and recorded.

\begin{algorithm}[!t]
\small

 \KwResult{10kHz sampling data of pressure in reservoir ($P_{reservoir}$), pressure in both valve chambers ($P_{1},P_{2}$) and volume flowrate in the controlled air channel ($Q_{controlled}$).}
 $P_{target}$ $\leftarrow$ 35 kPa;  \\
 \For{trial number $\leftarrow$ 1 to N}{
  Valve S1 to 5 $\leftarrow$ OFF;\\
  set $P_{reservoir}$ to $P_{target}$ via PID;\\
  Valve S2 $\leftarrow$ ON ;\\ 
  Pause for 10 seconds;\\
  Valve S2 $\leftarrow$ OFF $\;$  Valve S1 $\leftarrow$ ON ;\\
  Pause for 5 seconds;\\
  Valve S1 $\leftarrow$ OFF ;\\
  set $P_{reservoir}$ to $P_{target}$ via PID;\\
  Valve S5 $\leftarrow$ ON ;\\ 
  Pause for 10 seconds;\\
  Valve S5 $\leftarrow$ OFF $\;$  Valve S4 $\leftarrow$ ON ;\\
  Pause for 5 seconds;\\
  Valve S4 $\leftarrow$ OFF ;\\
  Save $P_{reservoir}$,$P_{1},P_{2}$ and $Q_{controlled}$
 }
 \caption{Procedure for fatigue tests.}
\end{algorithm}

\section{Experimental results}

\subsection{Results on Valve Behaviour over different design parameters}
\label{sec_results_parameter}
\begin{figure*}[!t]
\begin{center}
  \includegraphics[width=0.9\textwidth]{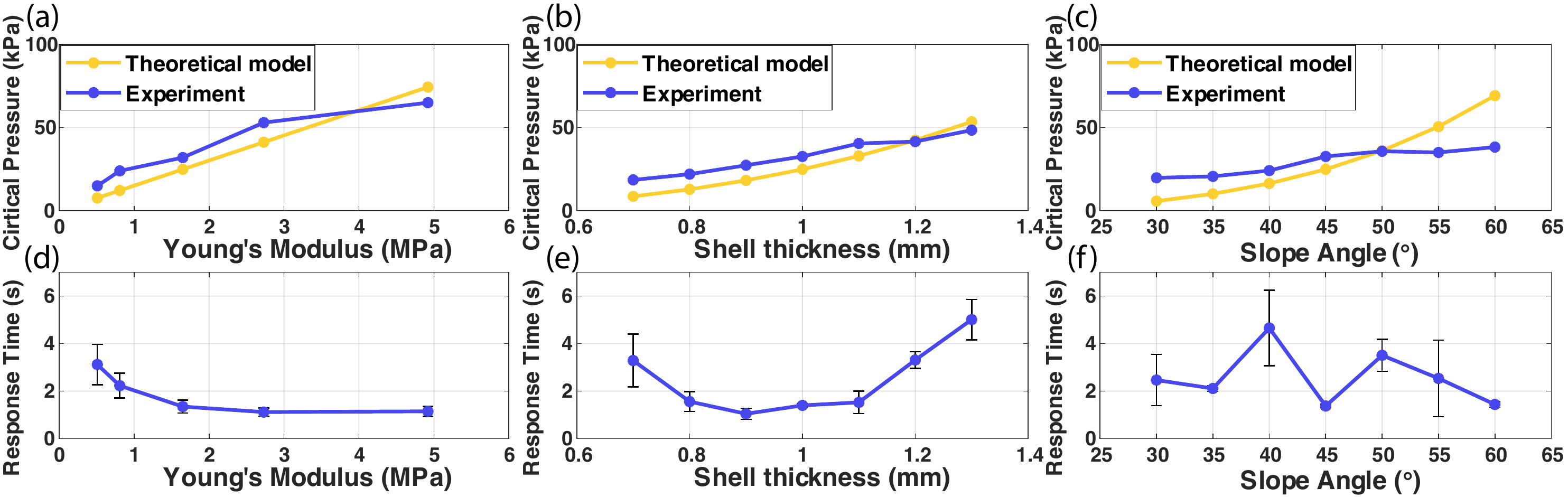}
  \vspace{-3mm} 
  \caption{\textcolor{black}{The measured critical pressure of the valves with different shell Young's Modulus \textbf{(a)}, shell thickness \textbf{(b)}, and shell slope angle values \textbf{(c)}. The prediction given by the theoretical model is plotted in the same graph for comparison. The measured response time of the valves with different shell material Young's Modulus \textbf{(d)}, shell thickness \textbf{(e)}, and shell slope angle values \textbf{(f)}. The error bar indicates the standard deviation over three repeated trials of the same valve. The switching direction during measurement is randomly chosen.} }
  \label{Fig_parameters}
  \vspace{-5mm}
  \end{center}
\end{figure*}

The effects of the design features on the critical pressure and response times are shown in Fig.~\ref{Fig_parameters}. Overall, the experiment results show a similar trend with the theoretical model for material hardness, membrane thickness, and slope angle with an RMSE of 9.66, 7.68 kPa and 15.2 kPa, respectively. The Young's modulus of each digital material used in the theoretical modelling is obtained from \cite{slesarenko2018towards,liu2018failure}. The theoretical model tends to underestimate the critical pressure when the critical pressure is relatively low, which may be caused by the frictional force exerted on the moving piston. Changing the material shore hardness achieves the largest adjustable critical pressure range (from 15.3 to 65.2 kPa with a maximum-minimum ratio of 4.26) among the three features. In contrast, changing the shell thickness can adjust the critical pressure from 18.6 to 48.5 kPa (a maximum-minimum ratio of 2.61) and hanging the slope angle can adjust the critical pressure from 19.8 to 38.3 kPa (a maximum-minimum ratio of 1.93).

\textcolor{black}{Fig.~\ref{Fig_parameters}(d) indicates that the response time decreases as the material Young's Modulus increases.} When tuning the shell thickness, the response time reaches its minimum value at around 1mm. No clear trend was found between the response time and the slope angle. \textcolor{black}{Although all three design dimensions successfully adjusted the critical switching pressure, they come with different performance in terms of consistency.} Changing the slope angle brings a large RMSE (15.2 kPa) between the predicted and experimental results. It also brings a large standard deviation (0.752 s) on response time (shown in Fig.~\ref{Fig_parameters} (d),(e),(f)) compared to an average of 0.404 s STD for material stiffness change and 0.499 s STD for membrane thickness change. Part of the reasons for its poor consistency is that changing the slope angle brings significant change to the overall geometry and size of the valve. We recommend using material hardness to adjust the critical pressure to suit different needs, as they bring a smaller RMSE between the theoretical model and the experiment results. Tuning the material hardness can also reduce the valve response time, while Fig.~\ref{Fig_parameters}(d) shows the fastest average response time ($\leq 1.8 s$). \textcolor{black}{The response time of the valve is considerably larger than the response time of the \textcolor{black}{silicone} cast soft fluidic valve (0.4 s) \cite{rothemund2018soft} or conventional solenoid valve (30-200 ms).}

\subsection{Results on Valve Behaviour and Consistency over different post-processing method}\label{sec_repeatability_result}

\begin{figure}[!t]
\begin{center}
  \includegraphics[width=\columnwidth]{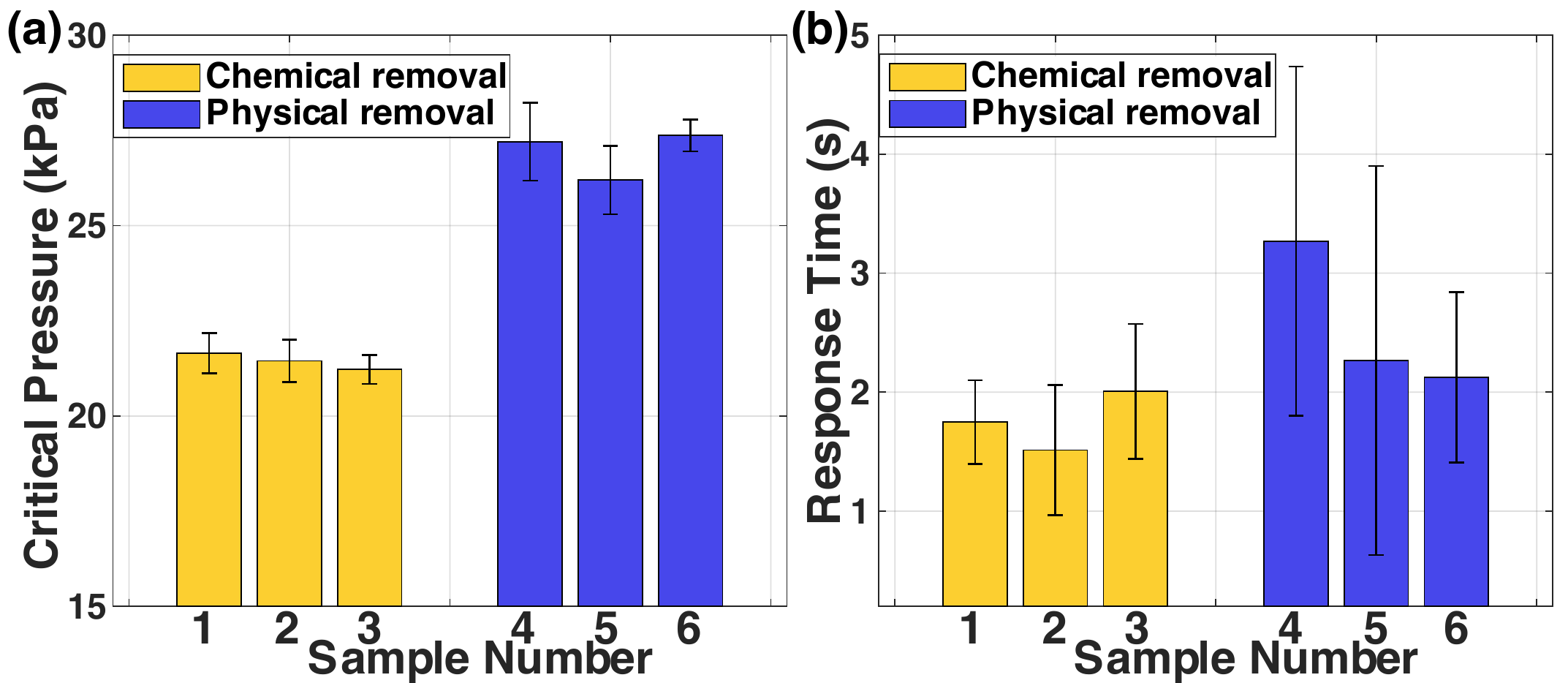}
  \vspace{-3mm} 
  \caption{The repeatability test of multiple valve samples with identical design parameters. Sample 1-3 are post-processed through 12 hrs chemical bathing, while sample 4-6 are post-processed through water-jetting. The \textbf{(a)} critical pressure and (\textbf{b}) response time of all six samples are presented. The error bar indicates the standard deviation of each sample over five trials. \textcolor{black}{The switching direction during measurement is randomly chosen.}}
  \label{repeatability}
  \vspace{-7mm}
  \end{center}
\end{figure}

The repeatability test shows a significant performance difference between valves with chemical and physical post-processing methods. As shown in Fig.~\ref{repeatability}, removing the support with chemical bathing decreased the critical switching pressure by $20.36\%$ and reduced response time by $31.23\%$. \textcolor{black}{The decrease in critical pressure after chemical support removal is due to some material property changes brought by the bathing process. It is also found that the samples with physical support removal technique come with poorer model consistency, especially in terms of the response time. The inconsistency is mainly caused by the tiny cracks on the surface of the compliant conical shells brought by the high pressure water jets. The generation of these cracks are highly unpredictable and irregular, therefore bringing a large inconsistency between different samples. The size of these cracks may also keep growing during each measurement, therefore bringing inconsistency between different trials even for the same sample. The large deviation in response time of these valves could bring design constraints and problems when controlling the soft robots which require constant-frequency periodic actuation or fast response to the environment.}


\subsection{Results on Fatigue Tests over different post-processing method}\label{sec_fatigue_results}

\begin{figure}[!t]
\begin{center}
  \includegraphics[width=\columnwidth]{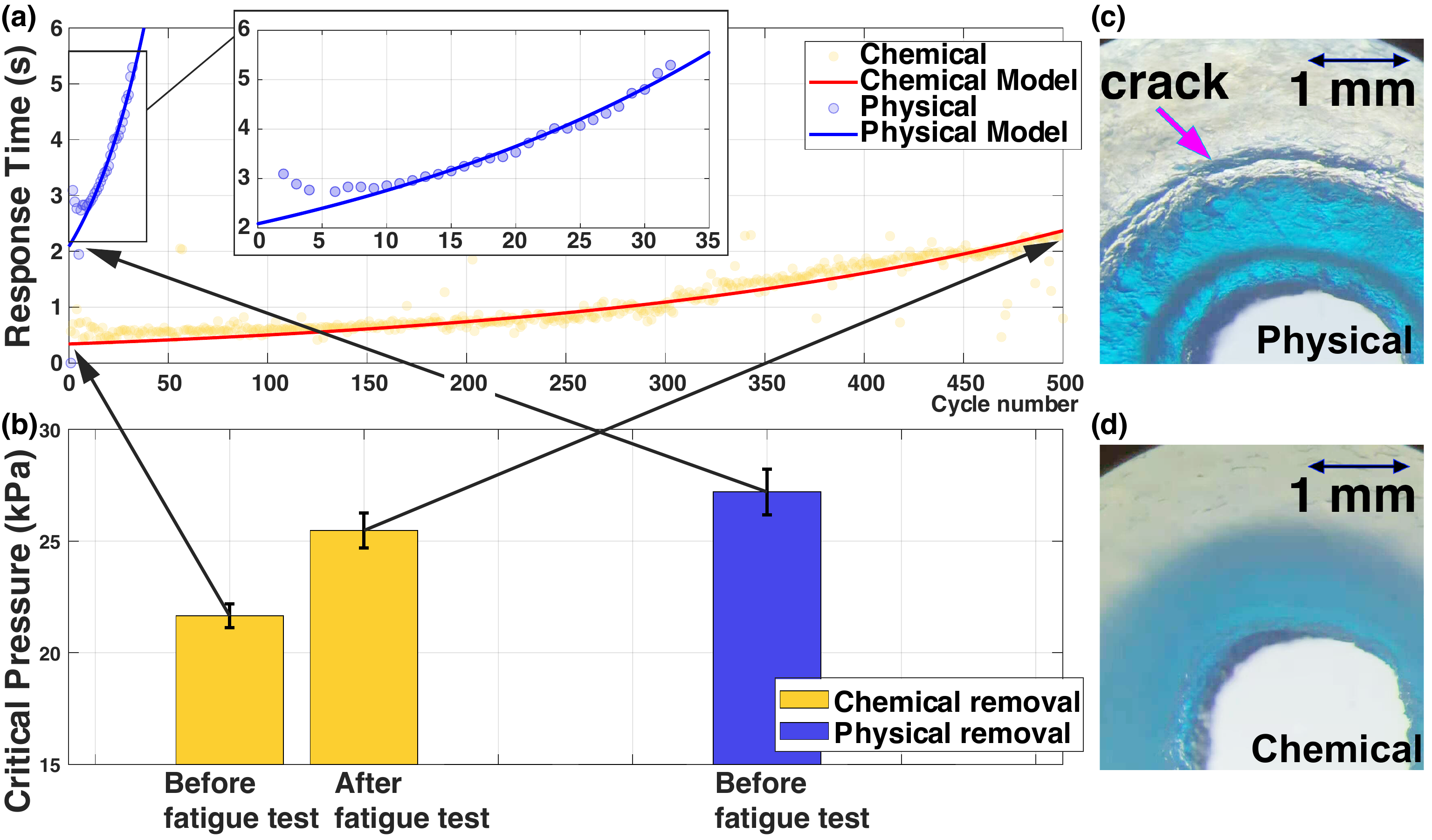}
  \vspace{-3mm} 
  \caption{Fatigue tests for valves whose support materials were removed by chemical bathing or physical method. \textbf{(a)} Measured response times at each trial. Note that the response time here is measured when the driving pressure is 35 kPa. Exponential models are used to fit the empirical data. (\textbf{b}) Measured critical switching pressure and corresponding response time before and after the fatigue tests. The error bar indicates the standard deviation of each sample over five trials. \textcolor{black}{After five cycles of fatigue test, small cracks on the sample post-processed with physical support removal can be observed in comparison to the sample post-processed with chemical bathing, shown in \textbf{c} and \textbf{d}, respectively. Both figures were captured using an optical microscope.}}
  \label{fatigue}
  \vspace{-7mm} 
  \end{center}
\end{figure}

The critical pressure increases with the increase of trial numbers during the fatigue test. Better performance of samples post-processed with chemical bathing is found with less critical pressure change and a pass of the fatigue test. \textcolor{black}{The sample post-processed with physical support material removal failed earlier after 32 complete cycles. This shorter life cycle is caused by the tiny cracks generated by the high-pressure water jets on the surface of the compliant conical shell (see Fig.~\ref{fatigue}(c) and (d)). These cracks gradually increase their size during the loading cycles and eventually breaks the shell and results in a leakage.}

An exponential model trained from the experiment data is used to further characterize the fatigue behaviour of the valve, as shown in Fig.~\ref{fatigue}. For the valve which uses chemical support-removal technique and physical support-removal technique, the fitted model are:
\begin{equation}
    T = 0.341 \times e^{0.00388\cdot n}
\end{equation}
and
\begin{equation}
        T = 2.09 \times e^{0.0280\cdot n}
\end{equation}
where $T$ is the response time and $n$ is the number of working cycles undertaken by the valve.

\section{Conclusion}\label{sec_conclusion}
This paper presents a 3D printed bistable pneumatic valve design with a method to tune the critical pressure with predefined design parameters. Inspired by the structural buckling of two conical bistable shells, the design utilizes multi-material 3D printing technology to create an electronics-free valve that embeds a pneumatic signal as input to control the state of connected air channels. A theoretical model was included to investigate the bistability and snap-through behaviour of the valve. The two local valleys observed on the strain energy curve reflect the bi-stable nature of the structure. The bistability of the valve ensures that no constant energy supply is required to maintain the state of the valve. Meanwhile, the pressure-displacement curve predicted the critical switching pressure required to switch the valve state. 

The experiment characterization further verifies the theoretical modelling results, showing that the critical switching pressure of the valve can be easily tuned within the range from 15.3 kPa to 65.2 kPa by adjusting the thickness, the slope angle and the material hardness of the conical shell in the valve design. This design flexibility and fast fabrication solution with 3D printing indicate the valve can be used as a digital valve that takes digital pneumatic signal as input and a comparator that deals with analogue pneumatic signal computation. \textcolor{black}{The light-weight and electronics-free nature of the valve allows it to be embedded in untethered soft robots, even in challenging environments where a conventional electronic system cannot be used.}  Thanks to the advantages of soft material 3D printing, the valve requires minimal manual work during its fabrication and component integration. The valve can be used to build logical gates, analogue-to-digital converters, and digital memory in future work. Although the critical switching pressure of the valve is characterized and presented by the design dimensions, the response time of the valve still requires more investigation. Compared with traditional mechanical valves, the 3D printed soft membrane (Agilus30) used in this work introduces higher viscous losses, increasing the response time. The valve also exhibits performance change after a large number of working cycles. In future studies, solutions that can increase the durability and reduce the response time of the valve will also be investigated.

\printbibliography

@article{slesarenko2018towards,
  title={Towards mechanical characterization of soft digital materials for multimaterial 3D-printing},
  author={Slesarenko, Viacheslav and Rudykh, Stephan},
  journal={International Journal of Engineering Science},
  volume={123},
  pages={62--72},
  year={2018},
  publisher={Elsevier}
}

@article{liu2018failure,
  title={Failure mechanism transition of 3D-printed biomimetic sutures},
  author={Liu, Lei and Li, Yaning},
  journal={Engineering Fracture Mechanics},
  volume={199},
  pages={372--379},
  year={2018},
  publisher={Elsevier}
}

@article{lee2017soft,
  title={Soft robot review},
  author={Lee, Chiwon and Kim, Myungjoon and Kim, Yoon Jae and Hong, Nhayoung and Ryu, Seungwan and Kim, H Jin and Kim, Sungwan},
  journal={International Journal of Control, Automation and Systems},
  volume={15},
  number={1},
  pages={3--15},
  year={2017},
  publisher={Springer}
}

@inproceedings{walker2020soft,
  title={Soft robotics: a review of recent developments of pneumatic soft actuators},
  author={Walker, James and Zidek, Thomas and Harbel, Cory and Yoon, Sanghyun and Strickland, F Sterling and Kumar, Srinivas and Shin, Minchul},
  booktitle={Actuators},
  volume={9},
  number={1},
  pages={3},
  year={2020},
  organization={Multidisciplinary Digital Publishing Institute}
}

@article{tang2020leveraging,
  title={Leveraging elastic instabilities for amplified performance: Spine-inspired high-speed and high-force soft robots},
  author={Tang, Yichao and Chi, Yinding and Sun, Jiefeng and Huang, Tzu-Hao and Maghsoudi, Omid H and Spence, Andrew and Zhao, Jianguo and Su, Hao and Yin, Jie},
  journal={Science Advances},
  volume={6},
  number={19},
  pages={eaaz6912},
  year={2020},
  publisher={American Association for the Advancement of Science}
}

@article{he2020soft,
  title={Soft fingertips with tactile sensing and active deformation for robust grasping of delicate objects},
  author={He, Liang and Lu, Qiujie and Abad, Sara-Adela and Rojas, Nicolas and Nanayakkara, Thrishantha},
  journal={IEEE Robotics and Automation Letters},
  volume={5},
  number={2},
  pages={2714--2721},
  year={2020},
  publisher={IEEE}
}

@article{rothemund2018soft,
  title={A soft, bistable valve for autonomous control of soft actuators},
  author={Rothemund, Philipp and Ainla, Alar and Belding, Lee and Preston, Daniel J and Kurihara, Sarah and Suo, Zhigang and Whitesides, George M},
  journal={Science Robotics},
  volume={3},
  number={16},
  year={2018},
  publisher={Science Robotics}
}

@article{gul20183d,
  title={3D printing for soft robotics--a review},
  author={Gul, Jahan Zeb and Sajid, Memoon and Rehman, Muhammad Muqeet and Siddiqui, Ghayas Uddin and Shah, Imran and Kim, Kyung-Hwan and Lee, Jae-Wook and Choi, Kyung Hyun},
  journal={Science and technology of advanced materials},
  volume={19},
  number={1},
  pages={243--262},
  year={2018},
  publisher={Taylor \& Francis}
}

@article{drotman2021electronics,
  title={Electronics-free pneumatic circuits for controlling soft-legged robots},
  author={Drotman, Dylan and Jadhav, Saurabh and Sharp, David and Chan, Christian and Tolley, Michael T},
  journal={Science Robotics},
  volume={6},
  number={51},
  year={2021},
  publisher={Science Robotics}
}

@ARTICLE{9044752,
  author={K. {Xu} and N. O. {Pérez-Arancibia}},
  journal={IEEE Robotics and Automation Letters}, 
  title={Electronics-Free Logic Circuits for Localized Feedback Control of Multi-Actuator Soft Robots}, 
  year={2020},
  volume={5},
  number={3},
  pages={3990-3997}
}

@article{luo2019soft,
  title={Soft kink valves},
  author={Luo, Kai and Rothemund, Philipp and Whitesides, George M and Suo, Zhigang},
  journal={Journal of the Mechanics and Physics of Solids},
  volume={131},
  pages={230--239},
  year={2019},
  publisher={Elsevier}
}

@book{megson2019structural,
  title={Structural and stress analysis},
  author={Megson, Thomas Henry Gordon},
  year={2019},
  publisher={Butterworth-Heinemann}
}

@article{tanaka2021dynamic,
  title={Dynamic turning of a soft quadruped robot by changing phase difference},
  author={Tanaka, Hiroaki and Chen, Tsung-Yuan and Hosoda, Koh},
  journal={Frontiers in Robotics and AI},
  volume={8},
  pages={93},
  year={2021},
  publisher={Frontiers}
}

@article{he2021method,
  title={A method to 3D print a programmable continuum actuator with single material using internal constraint},
  author={He, Liang and Tan, Xinyang and Suzumori, Koichi and Nanayakkara, Thrishantha},
  journal={Sensors and Actuators A: Physical},
  volume={324},
  pages={112674},
  year={2021},
  publisher={Elsevier}
}

@article{akbari2018enhanced,
  title={Enhanced multimaterial 4D printing with active hinges},
  author={Akbari, Saeed and Sakhaei, Amir Hosein and Kowsari, Kavin and Yang, Bill and Serjouei, Ahmad and Yuanfang, Zhang and Ge, Qi},
  journal={Smart Materials and Structures},
  volume={27},
  number={6},
  pages={065027},
  year={2018},
  publisher={IOP Publishing}
}

\end{document}